\documentclass[10pt,twocolumn,letterpaper]{article}

\usepackage{iccv}
\usepackage{times}
\usepackage{epsfig}
\usepackage{amsmath,amsfonts}
\usepackage{algorithmic}
\usepackage{algorithm}
\usepackage{array}
\usepackage[caption=false,font=normalsize,labelfont=sf,textfont=sf]{subfig}
\usepackage{textcomp}
\usepackage{stfloats}
\usepackage{url}
\usepackage{verbatim}
\usepackage{graphicx}
\usepackage{cite}
\usepackage{amssymb}
\usepackage{bbm}
\usepackage{color}
\usepackage{multirow}
\usepackage{bigstrut}
\usepackage[figuresright]{rotating}
\usepackage{makecell}
\usepackage{array}
\usepackage{float}
\usepackage[accsupp]{axessibility}

\usepackage[breaklinks=true,bookmarks=false]{hyperref}

\iccvfinalcopy 

\def\iccvPaperID{5076} 

\ificcvfinal\fi

\begin{document}

\title{Modality Unifying Network for Visible-Infrared Person Re-Identification}
\author{Hao Yu\textsuperscript{1},\quad Xu Cheng\textsuperscript{1}\thanks{Corresponding Author (Email: xcheng@nuist.edu.cn)},\quad Wei Peng\textsuperscript{2},\quad Weihao Liu\textsuperscript{3},\quad Guoying Zhao\textsuperscript{4}\\
\textsuperscript{1}School of Computer Science, Nanjing University of Information Science and Technology, China\\
\textsuperscript{2}Department of Psychiatry and Behavioral Sciences, Stanford University, USA\\
\textsuperscript{3}School of Computer Science and Technology, Soochow University, China\\
\textsuperscript{4}Center for Machine Vision and Signal Analysis, University of Oulu, Finland\\
{\tt\small \{yuhao,xcheng\}@nuist.edu.cn, wepeng@stanford.edu, whliu@stu.suda.edu.cn, guoying.zhao@oulu.fi}}
\maketitle
\ificcvfinal\thispagestyle{empty}\fi

\begin{abstract}
    Visible-infrared person re-identification (VI-ReID) is a challenging task due to large cross-modality discrepancies and intra-class variations. Existing methods mainly focus on learning modality-shared representations by embedding different modalities into the same feature space. As a result, the learned feature emphasizes the common patterns across modalities while suppressing modality-specific and identity-aware information that is valuable for Re-ID. To address these issues, we propose a novel Modality Unifying Network (MUN) to explore a robust auxiliary modality for VI-ReID. First, the auxiliary modality is generated by combining the proposed cross-modality learner and intra-modality learner, which can dynamically model the modality-specific and modality-shared representations to alleviate both cross-modality and intra-modality variations. Second, by aligning identity centres across the three modalities, an identity alignment loss function is proposed to discover the discriminative feature representations. Third, a modality alignment loss is introduced to consistently reduce the distribution distance of visible and infrared images by modality prototype modeling. Extensive experiments on multiple public datasets demonstrate that the proposed method surpasses the current state-of-the-art methods by a significant margin.
\end{abstract}

\section{Introduction}

Person re-identification (Re-ID) \cite{leng2019survey,ye2021deep} aims at matching pedestrian images captured from multiple non-overlapping cameras. Over the past few years, it has received increased attention due to its huge practical value in modern surveillance systems. Previous studies \cite{ning2020feature,yan2021beyond,zhu2020aware,liao2022graph,luo2019strong} mainly focus on matching pedestrian images captured from visible cameras and formulate the Re-ID task as a single-modality matching issue. Nevertheless, visible cameras may not provide accurate appearance information about persons in scenarios with poor illumination. To address this limitation, modern surveillance systems also employ infrared cameras, which can capture clear images in low-light conditions at night. As a result, visible-infrared person re-identification (VI-ReID) \cite{wu2017rgb,wu2021discover,chen2022structure} has become a topic of growing interest in recent times, which seeks to match infrared images of the same identity when given a visible query across multiple camera views and vice versa.

VI-ReID is challenging due to the huge cross-modality discrepancy between visible and infrared images, as well as the intra-modality variation in person bodies such as pose variation and dress change. Existing methods \cite{wu2021discover,ye2021channel,park2021learning,zhang2022cross,chen2022structure,zhang2022fmcnet} primarily focus on relieving the cross-modality discrepancy by extracting modality-shared features to perform the feature-level alignment. Some studies \cite{wu2017rgb,park2021learning,ye2021channel,ye2018visible,chen2022structure} employ two-stream networks for cross-modality feature embedding, while others \cite{wang2020cross,zhang2022fmcnet,dai2018cross,wang2019rgb} utilize Generative Adversarial Networks (GANs) to generate shared representations from visible and infrared images. However, these methods discard modality-specific features (such as colour and texture) that contain useful identity-aware patterns against intra-modality variations. Consequently, the learned features may not fully capture the variation of human bodies and thus lack discriminability. To address this limitation, the modality-unifying methods, \eg, X-modality \cite{li2020infrared}, DFM \cite{kong2021dynamic}, SMCL \cite{wei2021syncretic}, have been proposed to acquire the auxiliary modality by fusing visible and infrared modalities, encoding both modality-specific and modality-shared patterns to jointly relieve cross- and intra-modality discrepancies. In the SMCL \cite{wei2021syncretic}, the authors proposed a syncretic modality generated by fusing visible and infrared pixels, which can bridge the gap between visible and infrared modalities while maintaining discriminability as the modality-specific information is preserved.

However, the existing modality-unifying works still have three weaknesses. (1) \textbf{Pixel fusion.} Previous works generate the auxiliary modality by fusing pixels of the raw visible and infrared images, which makes the richness of semantic patterns either equal to the original two modalities or lower in the case of pixel misalignment. In fact, the auxiliary modality is utilized to guide the learning of visible and infrared modalities, but the insufficient semantic patterns lead to a lack of identity-related information and severely limit the capacity to relieve intra-modality variations in VI-ReID.\\ 
(2) \textbf{Discrepancy bias.} During the VI training, the relative distances between visible and infrared images are constantly changing, which causes a dynamic bias toward the balance of intra- and cross-modality discrepancies. Thus, an ideal auxiliary modality should be able to dynamically control the ratio of modality-specific and modality-shared patterns it contains to model the evolving modality discrepancies. However, the existing studies simply use the global information of visible and infrared images to obtain the auxiliary representations, which are inflexible in adjusting the patterns they describe, leading to low robustness.\\ 
(3) \textbf{Inconsistency constraints.} Existing studies usually utilize features in the current batch to represent the overall distribution for distance optimization. However, this strategy suffers from randomness, as the training samples are different in each batch, which may cause a certain inconsistency in the learned feature relationships in different training stages, thus damaging the generalizability.

\begin{figure}[t]
\centering
\includegraphics[width=3.2in]{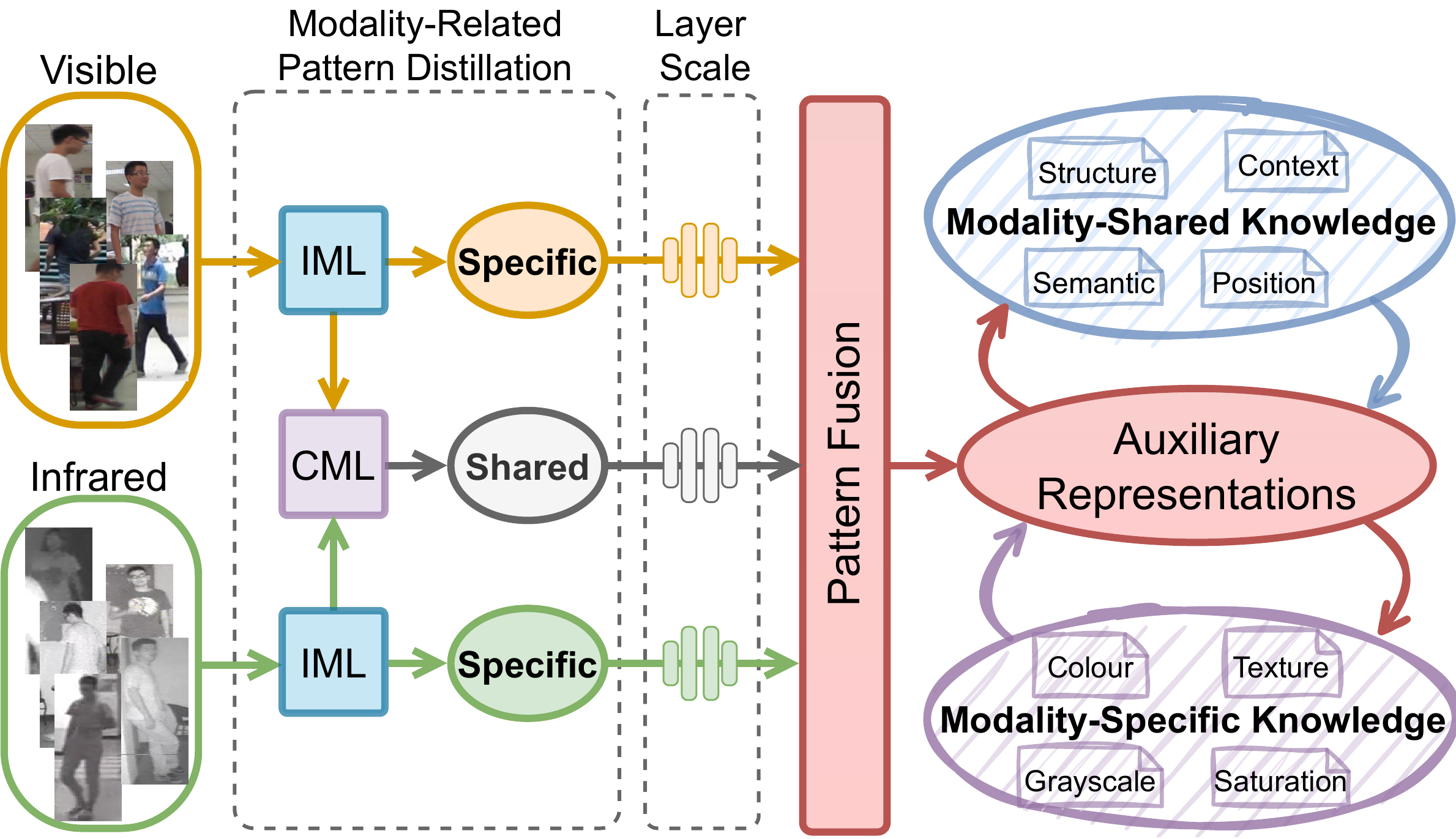}
\setlength{\abovecaptionskip}{0cm}
\caption{The main idea behind generating a strong auxiliary modality for the VI-ReID task. The IML and CML denote the intra-modality learner and cross-modality learner, respectively.}
\label{fig:1}
\vspace{-0.5em}
\end{figure}

Inspired by the above discussions, we propose a novel Modality Unifying Network (MUN) to explore an effective and robust auxiliary modality for VI-ReID. The main idea of our auxiliary modality is illustrated in Figure \ref{fig:1}. Specifically, we introduce an auxiliary generator comprising two intra-modality learners (IML) and one cross-modality learner (CML) to distil the modality-related patterns from visible and infrared images. Two IMLs are presented to identify the modality-specific and identity-aware patterns from visible and infrared images, respectively. They exploit multiple depth-wise convolutions with various kernel sizes to capture fine-grained semantic patterns in the human body across multiple receptive fields. Based on the outputs of two IMLs, the CML leverages spatial pyramid pooling to extract multi-scale feature representations and then fuse the modality-shared patterns learned in each feature scale. By combining IML and CML, the proposed auxiliary generator can generate a powerful auxiliary modality that is rich in modality-shared and discriminative patterns to alleviate both cross-modality and intra-modality discrepancies. In addition, the layer scale scheme is used to control the ratio of patterns learned from IML and CML, which can dynamically adjust the modality-specific and modality-shared patterns in the generated auxiliary representation. 

Furthermore, to reveal the identity-related patterns in each identity set, an effective identity alignment loss ($L_{ia}$) is designed to optimize the distances of tri-modality identity centres. In addition, to regulate the distribution level feature relationships while relieving the inconsistency issue caused by sample variations, a novel modality alignment loss ($L_{ma}$) is designed to minimize the distances of three modalities, which utilizes the modality prototype to represent the learned modality information in each iteration.

In general, the major contributions of this paper can be summarized as follows.
\begin{itemize}
    \item We propose a novel modality unifying network for the VI-ReID task by constructing a robust auxiliary modality, which contains rich semantic information from visible and infrared images to address modality discrepancies and reveal discriminative knowledge.
    \item A novel auxiliary generator constructed by the intra-modality and cross-modality learners is introduced to dynamically extract identity-aware and modality-shared patterns from heterogeneous images.
    \item The identity alignment loss and modality alignment loss are designed to jointly explore the generalized and discriminative feature relationships of the three modalities at both the identity and distribution levels.
    \item Extensive experiments on several public VI-ReID datasets verify the effectiveness of the proposed method and modality unifying scheme, which outperforms the current state of the arts by a large margin.
\end{itemize}


\begin{figure*}[t]
\centering
\includegraphics[width=0.92\linewidth]{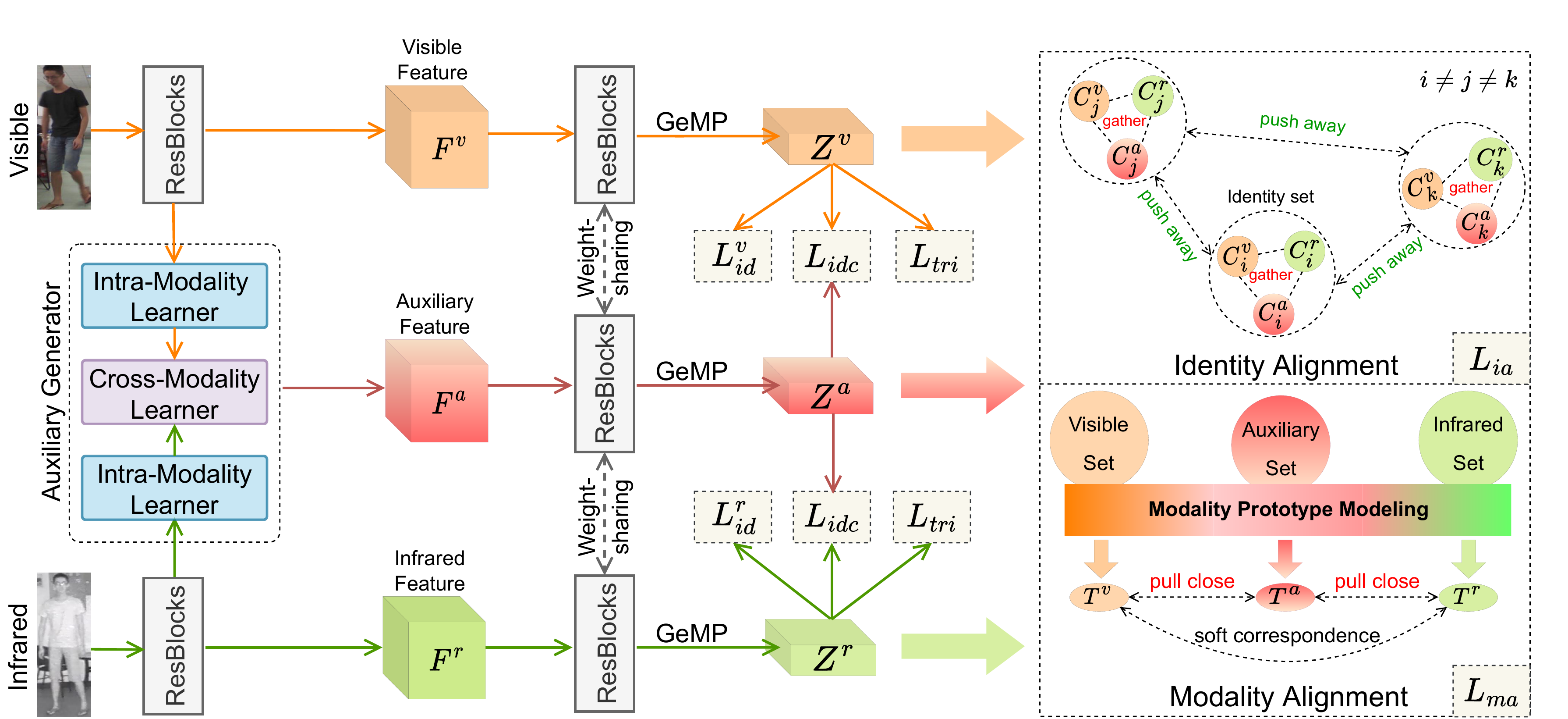}
\caption{The overall architecture of the proposed MUN for VI-ReID. GeMP denotes the Generalized Mean Pooling \cite{radenovic2018fine}. The pretrained ResNet-50 \cite{he2016deep} is adopted as the baseline network. To meet the specific requirements of VI-ReID, we initialize the first stage of the ResNet-50 twice as two independent ResBlocks to extract the low-level visible and infrared features, respectively. The remaining stages are utilized as modality-shared ResBlocks. During the inference, only visible and infrared modalities are utilized to perform cross-modality retrieval.}
\label{fig:2}
\end{figure*}

\section{Related Work}

\textbf{Single-Modality Person Re-ID.} Single-modality person re-identification \cite{ye2021deep,mathur2020brief,liu2022survey} aims to match pedestrian images across different visible cameras. It presents challenges such as changes in viewpoint and human pose across camera views. Current approaches mainly focus on feature representation learning \cite{ning2020feature,zheng2019joint,luo2019bag} and distance metric learning \cite{zhu2020aware,yan2021beyond,zeng2020hierarchical,liao2022graph}. Over the past few years, excellent performances have been achieved on several academic benchmarks. However, in practical scenarios, numerous crucial surveillance photos and videos are captured at night using infrared cameras. When it comes to matching pedestrians across visible and infrared modalities, the capabilities of these single-modality methods are limited due to their inability to address the huge modality gap. In contrast, we present an effective modality unifying network to bridge the modality gap and achieve precise cross-modality pedestrian matching in 24-hour monitoring scenarios.

\textbf{Visible-Infrared Person Re-ID.} Visible-Infrared person Re-ID \cite{wu2017rgb} is a challenging task due to the cross-modality discrepancies between visible and infrared images, as well as the intra-modality variations such as pose and dress changes. Existing studies \cite{wu2021discover,ye2018visible,ye2020dynamic,ye2021channel,park2021learning} mainly focus on learning the modality-shared representations to align the visible and infrared modalities. Some generation-based methods \cite{wang2020cross,zhang2022fmcnet,wang2019learning,wang2019rgb} have been developed to achieve modality alignment or translation by using Generative Adversarial Network (GAN). For instance, Wang \etal \cite{wang2019rgb} proposed a dual-alignment network that used GAN to jointly learn pixel and feature level alignment. The D2RL \cite{wang2019learning} is proposed to perform image-level modality translation by adversarial training that relieves the cross-modality discrepancy. Other works \cite{wu2021discover,ye2021channel,park2021learning,chen2022structure,ye2018visible} attempt to learn modality-shared features by designing two-stream networks to perform cross-modality feature embedding. Ye \etal \cite{ye2018visible} proposed a dual-constrained top-ranking method with a weight-shared two-stream network. Wu \etal \cite{wu2021discover} designed a cross-modality attention scheme to help the two-stream backbone discover cross-modality nuances. However, these methods usually discard modality-specific representations that help to relieve intra-modality variations, leading to low robustness and discriminability in learned features.

In order to capture both the modality-shared and identity-aware patterns from heterogeneous images, modality-unifying methods have been developed. These methods aim to obtain the auxiliary modality by combining modality-specific and modality-shared representations from both visible and infrared images. The syncretic modality \cite{wei2021syncretic} is proposed to guide the generation of discriminative and modality-invariant representations. The DFM \cite{kong2021dynamic} acquires the mixed modality by integrating visible and infrared pixels. However, these methods generate the auxiliary modality by directly fusing the raw pixels of visible and infrared images, making their auxiliary modality lack high-level semantic patterns and inflexible to adjust its representations. 

To tackle these challenges, this paper presents the intra-modality learner and cross-modality learner to dynamically uncover substantial modality-shared and discriminative patterns from multiple receptive fields and feature scales. By integrating these learners, we introduce a powerful auxiliary modality that effectively bridges the modality discrepancy and enhances the discriminability of learned features.

\section{Methodology}
As shown in Figure \ref{fig:2}, we introduce the details of the Modality Unifying Network. We first utilize two independent ResBlocks to extract low-level features from visible and infrared images, respectively. Then, the auxiliary generator is designed to generate the auxiliary features by combining intra-modality and cross-modality learners. Afterwards, the visible, infrared, and auxiliary features are fed into weight-shared ResBlocks to learn high-level patterns. The auxiliary features can serve as a bridge to relieve both intra- and cross-modality discrepancies during the training. Based on the visible, infrared, and auxiliary features learned by weight-shared ResBlocks, four loss functions are developed to effectively improve cross-modality matching accuracy, including identity loss $L_{id}$, identity consistency loss $L_{idc}$, identity alignment loss $L_{ia}$ and modality alignment loss $L_{ma}$.

\subsection{Auxiliary Generator}
\begin{figure}[t]
\centering
\includegraphics[width=0.95\linewidth]{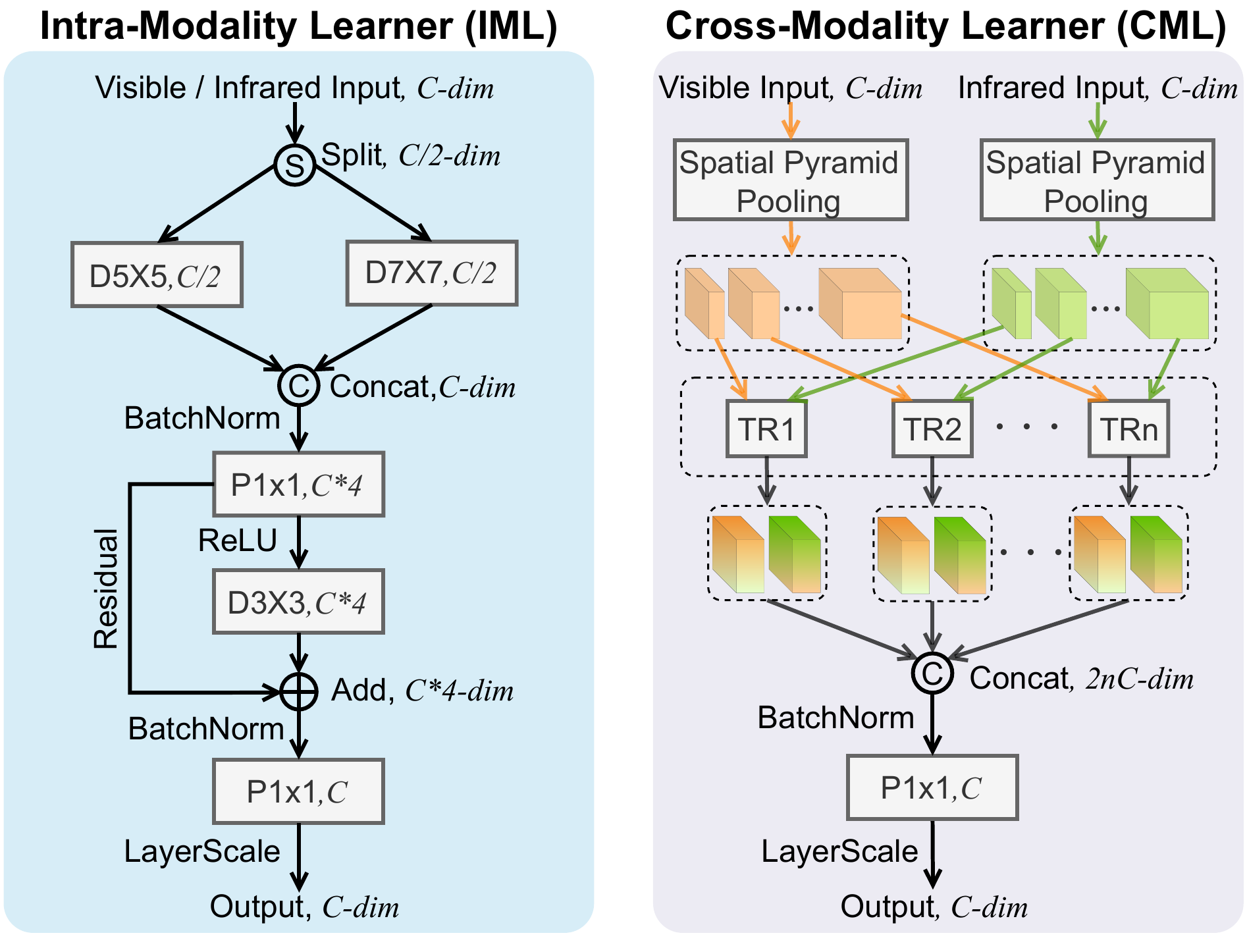}
\caption{The detailed architecture of the proposed intra-modality learner (IML) and cross-modality learner (CML). They are designed to decouple the modeling of modality-related knowledge.}
\label{fig:3}
\end{figure}

The auxiliary generator contains two intra-modality learners (IML) and one cross-modality learner (CML). The two IMLs are designed to mine identity-related patterns from visible and infrared images, respectively. The CML is designed to learn modality-shared patterns based on the outcomes of two IMLs. The detailed architectures of IML and CML are shown in Figure \ref{fig:3}.

\textbf{Intra-Modality Learner.} The intra-modality learner (IML) is designed to capture the discriminative and identity-aware patterns in human bodies. The visible or infrared low-level features $\small{\textbf{F}^m \in \mathbb{R}^{C\times H\times W}, m\in \{v,r\}}$ extracted from two independent ResBlocks are regarded as the input of IML, where $m$ denotes the visible or infrared modality.

To enrich the receptive field while keeping low computational complexity, we equally divide $\small{\textbf{F}^m}$ into two parts along the channel dimension by matrix slice operation.
\begin{equation}
\begin{small}
\begin{aligned}
 \textbf{F}^{m}_{c_1} = \textbf{F}^{m}[0:C/2, :, :], \quad\textbf{F}^{m}_{c_2} = \textbf{F}^{m}[C/2:C, :, :].
\label{eq:1}
\end{aligned}
\end{small}
\end{equation}

Then, we employ $7\times7$ and $5\times5$ depth-wise convolutions ($D$) to operate on $\textbf{F}^{m}_{c_1}$ and $\textbf{F}^{m}_{c_2}$, respectively. This allows us to capture spatial patterns in different receptive fields.
\begin{equation}
\begin{small}
\begin{aligned}
 \textbf{R}^{m} = Concat\{D_{5\times5}(\textbf{F}^{m}_{c_1}), D_{7\times7}(\textbf{F}^{m}_{c_2})\},
\label{eq:2}
\end{aligned}
\end{small}
\end{equation}
where $Concat$ denotes the concatenation on channel dimension; $\textbf{R}^{m}$ indicates the visible or infrared features captured from multiple receptive fields. Then, a point-wise convolution ($P$) is utilized to fuse patterns with diverse receptive fields by connecting pixels in each channel.
\begin{equation}
\begin{small}
\begin{aligned}
 \textbf{R}^{m}_{1} = P_{1\times1}(BatchNorm(\textbf{R}^{m})).
\label{eq:3}
\end{aligned}
\end{small}
\end{equation}

To integrate and encode the structural information in human bodies, another depth-wise convolution with $3\times3$ kernel size is introduced to remodel the learned spatial map. This layer also utilizes a residual branch to retain information from the previous layer.
\begin{equation}
\begin{small}
\begin{aligned}
 \textbf{R}^{m}_{2} = D_{3\times3}(ReLU(\textbf{R}^{m}_{1})) + \textbf{R}^{m}_{1}.
\label{eq:4}
\end{aligned}
\end{small}
\end{equation}

In addition, another point-wise convolution is utilized to fuse patterns with diverse receptive fields in $\textbf{R}^{m}_{2}$.
\begin{equation}
\begin{small}
\begin{aligned}
 \hat{\textbf{F}}^{m} = I_{scale} *P_{1\times1}(BatchNorm(\textbf{R}^{m}_{2})),
\label{eq:5}
\end{aligned}
\end{small}
\end{equation}
where $I_{scale} \in(0,1]$ is the learnable layer scale factor used to control the ratio of intra-modality patterns learned by IML; $\hat{\textbf{F}}^{m}$ denotes the outcomes of the two IMLs.

In the proposed intra-modality learner, three depth-wise convolutions with various kernel sizes are well combined to capture the identity-related patterns that existed in various receptive fields. Two point-wise convolutions are utilized for pattern integration and channel relation reasoning based on the inverted residual architecture \cite{liu2022convnet}. The first point-wise convolution increases the channel dimension from $C$ to $C*4$ and the last point-wise convolution reduces the channel dimension from $C*4$ to $C$.

\textbf{Cross-Modality Learner.} The cross-modality learner is designed to mine the modality-shared patterns from multiple feature scales based on the outcomes of two IMLs. Specifically, the spatial pyramid features are mined by applying $n$ average pooling layers with various ratios.
\begin{equation}
\begin{small}
\begin{aligned}
 \textbf{S}^m_{1} &= Avgpool_1(\hat{\textbf{F}}^{m})&,\\
 \textbf{S}^m_{2} &= Avgpool_2(\hat{\textbf{F}}^{m})&,\\
&....&,\\
\textbf{S}^m_{n} &= Avgpool_n(\hat{\textbf{F}}^{m})&,
\label{eq:6}
\end{aligned}
\end{small}
\end{equation}
where $\{\textbf{S}^{m}_{1},\textbf{S}^{m}_{2},...,\textbf{S}^{m}_{n}\}, m\in\{v,r\}$ denote the spatial pyramid features with various feature scales. Afterwards, we obtain the modality-shared spatial patterns from each pair of cross-modality spatial pyramid features $\{\textbf{S}^{v}_{i},\textbf{S}^{r}_{i}\}_{i=1}^{n}$ with the same feature scale by using a group of learnable transposed convolutions $\{TR_1, TR_2,...,TR_n\}$.
\begin{equation}
\begin{small}
\begin{aligned}
 \hat{\textbf{S}}^v_{i} = TR_{i}({\textbf{S}}^v_{i}), \quad \hat{\textbf{S}}^r_{i} = TR_{i}({\textbf{S}}^r_{i}), \quad i &= 1,2,...,n.
\label{eq:7}
\end{aligned}
\end{small}
\end{equation}

Here, the spatial dimensions of each cross-modality feature pair $\small{\hat{\textbf{S}}^v_{i}}$ and $\small{\hat{\textbf{S}}^r_{i}}$ are rebuilt to $\small{H\times W}$ by the corresponding transposed convolution $TR_{i}$, respectively. In this manner, the significant patterns of visible and infrared features on each feature scale are embedded together, which helps to discover and amplify the abundant modality-shared information on multiple feature scales.

Further, all the embedded features are concatenated on the channel dimension via Eq.(8).
\begin{equation}
\begin{small}
\begin{aligned}
 \hat{\textbf{S}} = Concat\{\hat{\textbf{S}}^v_{i},\hat{\textbf{S}}^v_{2},...,\hat{\textbf{S}}^v_{n},\hat{\textbf{S}}^r_{1},\hat{\textbf{S}}^r_{2},...,\hat{\textbf{S}}^r_{n}\}.
\label{eq:8}
\end{aligned}
\end{small}
\end{equation}

Then, the auxiliary feature is obtained by fusing patterns captured from multiple feature scales. 
\begin{equation}
\begin{small}
\begin{aligned}
 \textbf{F}^{a} = C_{scale} * P_{1\times1}(BatchNorm(\hat{\textbf{S}})),
\label{eq:9}
\end{aligned}
\end{small}
\end{equation}
where $\textbf{F}^{a}$ denotes the auxiliary feature generated by our method. It contains substantial modality-shared and identity-aware information captured by three learners; $C_{scale}\in(0,1]$ denotes the learnable layer scale factor used to control the ratio of modality-shared representations in the learned auxiliary feature $\textbf{F}^{a}$; $P_{1\times1}$ is a point-wise convolution used to fuse patterns across different channels.

The CML mines significant patterns from multiple feature scales and amplifies the modality-shared parts from them using transposed convolutions. It makes our auxiliary feature a powerful tool to handle cross-modality variations.

\subsection{Classification Constraint}
To ensure the learned visible and infrared features are identity-related, the identity loss ($L_{id}$) implemented with the cross-entropy term is introduced as follows.
\begin{equation}
\begin{small}
\begin{aligned}
 L_{id}^{m} = &-\frac{1}{k}\sum_{i=1}^{k}logP(y_i|C_{m}(\textbf{Z}_{i}^{m})),\quad s.t.\quad m\in\{v,r\},
\label{eq:10}
\end{aligned}
\end{small}
\end{equation}
where $\textbf{Z}^v_i$ and $\textbf{Z}^r_i$ denote the generalized mean pooled visible and infrared features in the $i$-th identity, respectively. $k$ is the number of visible or infrared images in each batch; $y_i$ is the \emph{i}-th identity label; $C_{v}(\cdot)$ and $C_{r}(\cdot)$ are the predictions of visible and infrared classifiers, respectively.

The learned presentations are modality-shared if two classifiers can give consistent predictions for features from any modalities. However, if we directly apply features in one modality to the classifier in another modality  (\eg, $C_{r}(\textbf{Z}^{v})$), it may impose the classifier to learn modality-specific patterns rather than the modality-shared patterns, as the former is typically more discriminative. To solve this issue, we present an identity-consistency loss $L_{idc}$ to update the parameters of both visible and infrared classifiers with the aid of auxiliary features. It can be defined as follows.
\begin{equation}
\begin{small}
\begin{aligned}
 L_{idc} = -\frac{1}{k}\sum_{i=1}^{k}[logP(y_i|C_{v}(\textbf{Z}_{i}^{a}))+logP(y_i|C_{r}(\textbf{Z}_{i}^{a}))],
\label{eq:11}
\end{aligned}
\end{small}
\end{equation}
where $\textbf{Z}^{a}_i$ denotes the pooled auxiliary feature in the $i$-th identity. The auxiliary features effectively integrate visible and infrared patterns, facilitating the transfer of identity-related knowledge between modalities without compromising the original intra-modality learning.

\subsection{Identity Alignment Loss}
To relieve the class-level modality discrepancies and learn discriminative feature relationships, the identity alignment loss $L_{ia}$ is designed to align the visible and infrared features of each identity with the aid of auxiliary features.
\begin{equation}
\begin{small}
\begin{aligned}
 L_{ia} = \sum_{i=1}^{P}\Big[\alpha&+\mathop{max}\limits_{m_1\in\{v,r\}}|| \textbf{C}^a_i - \textbf{C}^{m_1}_i||_2 \\&- \mathop{min}\limits_{\mathop{m_2 \in \{v,r\}}\limits_{q \neq i}}||\textbf{C}^a_i - \textbf{C}^{m_2}_q||_2\Big],
\label{eq:12}
\end{aligned}
\end{small}
\end{equation}
where $\alpha$ is the margin parameter; $P$ denotes the number of person identities; $N$ is the number of images in the $i$-th identity; $\small{\textbf{C}^a_i = \frac{1}{N}\sum_{j=1}^{N}\textbf{Z}^{a}_{i,j}}$, $\small{\textbf{C}^v_i = \frac{1}{N}\sum_{j=1}^{N}\textbf{Z}^{v}_{i,j}}$, $\small{\textbf{C}^r_i = \frac{1}{N}\sum_{j=1}^{N}\textbf{Z}^{r}_{i,j}}$ are the auxiliary, visible, and infrared centres in the $i$-th identity, respectively; $\small{\textbf{Z}^v_{i,j}}$, $\small{\textbf{Z}^r_{i,j}}$ and $\small{\textbf{Z}^a_{i,j}}$ denote the $j$-th visible, infrared, and auxiliary features in the $i$-th identity set.

In this paper, identity alignment loss is proposed to optimize the hardest cross-modality positive and negative centre pairs in a triplet-metric manner. It regulates discriminative and robust feature relationships by forcing all identities to form a tight intra-class space and pushing centres of different identities away across the three modalities.

\subsection{Modality Alignment Loss}
Previous works \cite{wu2021discover,park2021learning,kong2021dynamic} typically align the two modalities by constraining visible and infrared features in each iteration. This scheme suffers from inconsistencies in the learned cross-modality feature relationships because the training samples are different in each iteration. To overcome this issue, we propose a modality alignment strategy that consistently aligns visible and infrared modalities by modeling the prototypes from features in each iteration.

Specifically, we first introduce three modality prototypes to represent the global information of visible, infrared, and auxiliary modalities, respectively. They can be denoted as $\textbf{T}^{v} = \{\textbf{t}_1^v,\textbf{t}_2^v,...,\textbf{t}_B^{v}\}$, $\textbf{T}^{r} = \{\textbf{t}_1^r,\textbf{t}_2^r,...,\textbf{t}_B^r\}$ and $\textbf{T}^{a} = \{\textbf{t}_1^a,\textbf{t}_2^a,...,\textbf{t}_B^a\}\in \mathbb{R}^{B \times C}$, where $\textbf{t}^v_i$, $\textbf{t}^r_i$ and $\textbf{t}^a_i$ are the modality prototypes for the $i$-th visible, infrared, and auxiliary features in each training batch ($\small{B}$), respectively. 

The initial prototypes of the three modalities are obtained based on the pooled features $[\textbf{Z}^{v}]^{0}$, $[\textbf{Z}^{r}]^{0}$ and $[\textbf{Z}^{a}]^{0}\in R^{B\times C}$ in the $0$-th iteration.
\begin{equation}
\begin{small}
\begin{aligned}
 [\textbf{T}^m]^{0} = \textbf{W}_p^{m}[\textbf{Z}^m]^{0}, \quad s.t.\quad m\in\{v,r,a\},
\label{eq:13}
\end{aligned}
\end{small}
\end{equation}
where $\textbf{W}_p^m$ are learnable matrices to distil modality-related patterns from the \emph{m}-th modality. $[\textbf{T}^m]^{0}$ denotes the prototype of the \emph{m}-th modality calculated in the $0$-th iteration. 

Further, to dynamically model the modality information during the training, we develop a temporal accumulation strategy to update the modality prototype by the learned features in each iteration, which can be defined as follows.
\begin{equation}
\begin{small}
\begin{aligned}
 [\textbf{T}^m]^{i} = [\beta]^{i}* \textbf{W}_p^{m}[\textbf{Z}^m]^{i}+(1-[\beta]^{i})*[\textbf{T}^{m}]^{i-1},\\
 s.t. \quad m\in\{v,r,a\},
\label{eq:14}
\end{aligned}
\end{small}
\end{equation}
where $[\textbf{T}^m]^{i}$ is the \emph{m}-th modality prototype calculated in the $i$-th iteration; $\beta$ is the updating ratio, which gradually increases from $1e^{-8}$ to 1 with the training goes on. The temporal accumulation strategy ensures that the modality information in each iteration is considered, thereby synchronizing the cross-modality alignment during the training.

Based on the modality prototypes, the modality alignment loss ($L_{ma}$) is designed as:
\begin{equation}
\begin{small}
\begin{aligned}
L_{ma}= \frac{1}{P}\sum_{p=1}^{P}[mmd(\textbf{T}^v_p, \textbf{T}^a_p) + mmd(\textbf{T}^a_p, \textbf{T}^r_p)],
\label{eq:15}
\end{aligned}
\end{small}
\end{equation}
where $\textbf{T}^v_p$, $\textbf{T}^r_p$ and $\textbf{T}^a_p$ denote visible, infrared, and auxiliary prototypes of the $p$-th identity, respectively. $mmd(\cdot,\cdot)$ is the MMD loss \cite{jambigi2021mmd} implemented the modality level alignment by constraining the distance of modality prototypes in each iteration. In Equation \ref{eq:15}, the $mmd(\textbf{T}^v_p, \textbf{T}^a_p)$ is defined as:
\begin{equation}
\begin{small}
\begin{aligned}
mmd(\textbf{T}^v_p,\textbf{T}^a_p)= || \frac{1}{N} \sum_{i=1}^{N} \phi(\textbf{T}^v_{p,i}) - \frac{1}{M} \sum_{j=1}^{M} \phi(\textbf{T}^a_{p,j})||^2_H,
\label{eq:16}
\end{aligned}
\end{small}
\end{equation}
where $\textbf{T}^v_{p,i}$ and $\textbf{T}^a_{p,j}$ denote the $i$-th visible prototype and the $j$-th auxiliary prototype in the $p$-th identity, respectively; $||\cdot||_H$ denotes the distribution measured by Gaussian kernel function $\phi(\cdot)$ which projects prototypes into the reproducing kernel Hilbert space. The $mmd(\textbf{T}^a_p, \textbf{T}^r_p)$ term can also be obtained in a similar way.
 
The modality alignment loss is used to constrain the identity-guided distribution distances of visible, infrared, and auxiliary modalities through the prototypes. It can effectively reduce the modality discrepancy and relieve the inconsistency issue in learned feature relationships. Meanwhile, the auxiliary modality can act as a bridge to decrease the relative distance between visible and infrared modalities in the common feature space, thereby significantly reducing the optimization difficulty of cross-modality alignment.

\subsection{Overall Loss Function}
Following the previous works, we employ the identity loss ($L_{id} = L_{id}^v+L_{id}^r$) and hard-mining triplet loss ($L_{tri}$) \cite{ye2020dynamic,wang2019learning} as our baseline loss functions. The overall loss function of the proposed MUN can be summarized as:
\begin{equation}
\begin{small}
\begin{aligned}
L_{total} = L_{id} + L_{tri} + \gamma*L_{idc} + \theta*L_{ia} + \sigma*L_{ma},
\label{eq:17}
\end{aligned}
\end{small}
\end{equation}
where $\gamma$, $\theta$, and $\sigma$ are parameters to balance the contribution of each proposed loss term during the training.

\section{Experiments}
\subsection{Datasets and Evaluation Settings}
\textbf{SYSU-MM01} \cite{wu2017rgb} is the largest dataset for VI-ReID, which comprises six cameras, including four visible and two infrared cameras. It encompasses a total of 491 individuals, with 287,628 visible images and 15,792 infrared images. The training set is composed of 395 individuals, with 22,258 visible images and 11,909 infrared images. The test set consists of 96 individuals, with 3,803 infrared images for queries and a gallery selected from 301 visible images. The dataset offers two testing settings: all-search mode and indoor-search mode. For both modes, we employ the hardest single-shot setting to perform the evaluation.

\textbf{RegDB} \cite{nguyen2017person} comprises 412 identities and a total of 8,240 images, with 206 identities allocated for training and another 206 identities for testing. Each identity is represented by 10 visible and 10 infrared images. The testing phase in RegDB involves two modes: Visible to Infrared, where visible images are searched using an infrared image, and Infrared to Visible, which entails the reverse scenario. For both modes, we repeat the testing process 10 times and average the results to report the mean values.

\textbf{Evaluation Settings.} We utilize the standard cumulative matching characteristics (CMC) and mean average precision(mAP) as the evaluation metrics.
\subsection{Implementation Details}
We implement all the experiments on the PyTorch framework with an NVIDIA RTX-3090 GPU. To ensure repeatability and facilitate fair comparisons with existing methods, we adopt the pretrained ResNet-50 \cite{he2016deep} as our backbone network, where the first stage is initialized twice as two modality-specific ResBlocks, and the rest stages are used as the modality-shared ResBlocks.

At the training stage, all images are resized to 288$\times$144. Data augmentations, including random horizontal flipping, erasing, and channel augmentations \cite{ye2021channel}, are utilized against overfitting. Our model is trained with the AdamW optimizer \cite{loshchilov2018fixing} for 90 epochs with a weight decay of 0.01. The learning rate is gradually increased from $10^{-8}$ to 0.002 in the first 15 epochs and then decays by 0.1 at the $30^{th}$ and $60^{th}$ epochs. We find the optimal settings of all the hyper-parameters by grid search and repeated ablation experiments. Specifically, the pooling ratios of the spatial pyramid pooling in CML are set to be \{2, 4, 6, 12\}; The margin parameter $\alpha$ is set to 0.55; the loss balance parameters $\gamma$, $\theta$, and $\sigma$ are set to 0.25, 0.5, and 0.008, respectively.
\subsection{Ablation Study}
We evaluate the effectiveness of each proposed component on SYSY-MM01 and RegDB datasets, as shown in Table \ref{tab:1}. Compared with the baseline (B) which only learns from visible and infrared modalities, the leveraging of auxiliary modality (Aux.) can effectively relieve both cross-modality and intra-modality discrepancies, thus greatly improving all the metrics on two datasets. Further, when applying the identity consistency loss ($L_{idc}$) to refine the modality-shared discriminative patterns, the performance is further improved. Meanwhile, the proposed identity alignment loss ($L_{ia}$) or modality alignment loss ($L_{ma}$) can enhance the cross-modality matching accuracy by aligning the visible and infrared features at the identity level or distribution level. By combining them, we can regulate a more robust cross-modality feature relationship, achieving a rank-1 of 76.24\% and mAP of 73.81\% on the SYSU-MM01 dataset. The results demonstrate that all the proposed components contribute consistently to the accuracy gain. 

It is worth noting that when adding only the auxiliary modality to the baseline in Table \ref{tab:1}, we employed a simple identity loss (like $L_{id}^a$) to supervise the auxiliary modality. This loss is deprecated when employing the proposed $L_{idc}$ to jointly supervise the three modalities.

\begin{table}[htbp]
  \centering
  \caption{Evaluation of different components of the proposed method on SYSU-MM01 and RegDB datasets. CMC (\%) at rank 1 and mAP (\%). The Red bold font and blue bold front denote the best and second best performances, respectively.}
  \resizebox{\linewidth}{!}{
    \begin{tabular}{ccccc|cc|cc}
    \hline
    \multirow{2}[4]{*}{B} & \multirow{2}[4]{*}{Aux.} & \multirow{2}[4]{*}{$L_{idc}$} & \multirow{2}[4]{*}{$L_{ia}$} & \multirow{2}[4]{*}{$L_{ma}$} & \multicolumn{2}{c|}{SYSU-MM01} & \multicolumn{2}{c}{RegDB} \bigstrut\\
\cline{6-9}          &       &       &       &       & r=1   & mAP   & r=1   & mAP \bigstrut\\
    \hline
    $\checkmark$ &       &       &       &       & 57.49 & 55.83 & 75.42 & 71.93 \bigstrut[t]\\
    $\checkmark$ & $\checkmark$ &       &       &       & 62.55 & 58.42 & 79.26 & 73.81 \\
    $\checkmark$ & $\checkmark$ & $\checkmark$ &       &       & 66.58 & 61.29 & 83.51 & 79.79 \\
    $\checkmark$ & $\checkmark$ & $\checkmark$ & $\checkmark$ &       & \color{blue}\textbf{71.35} & 65.54 & 89.42 & \color{blue}\textbf{84.66} \\
    $\checkmark$ & $\checkmark$ & $\checkmark$ &       & $\checkmark$ & 69.77 & \color{blue}\textbf{66.96} & \color{blue}\textbf{89.93} & 84.02 \\
    $\checkmark$ & $\checkmark$ & $\checkmark$ & $\checkmark$ & $\checkmark$ & \color{red}\textbf{76.24} & \color{red}\textbf{73.81} & \color{red}\textbf{95.19} & \color{red}\textbf{87.15} \\
    \hline
    \end{tabular}%
    }
  \label{tab:1}%
\end{table}%

\begin{table}[htbp]
  \centering
  \caption{Performance of using different intermediate modalities in our MUN on two datasets. CMC (\%) at rank 1 and mAP (\%).}
  \resizebox{\linewidth}{!}{
    \begin{tabular}{c|cc|cc}
    \hline
    \multirow{2}[4]{*}{Intermediate Modality} & \multicolumn{2}{c|}{SYSU-MM01} & \multicolumn{2}{c}{RegDB} \bigstrut\\
\cline{2-5}          & r=1   & mAP   & r=1   & mAP \bigstrut\\
    \hline
    X-modality \cite{li2020infrared} & 66.17 & 63.06 & 79.95 & 74.28 \bigstrut[t]\\
    Mixed modality \cite{kong2021dynamic} & 66.42 & 62.85 & 79.43 & 73.09 \\
    Syncretic modality \cite{wei2021syncretic} & \color{blue}\textbf{72.95} & \color{blue}\textbf{68.74} & \color{blue}\textbf{84.59} & \color{blue}\textbf{79.11} \\
    Auxiliary modality (Ours) & \color{red}\textbf{76.24} & \color{red}\textbf{73.81} & \color{red}\textbf{95.19} & \color{red}\textbf{87.15} \bigstrut[b]\\
    \hline
    \end{tabular}%
    }
  \label{tab:2}%
\end{table}%

\textbf{Effectiveness of auxiliary modality.} We conduct the ablations in the MUN by replacing our auxiliary modality with intermediate modalities designed by previous works, namely X \cite{li2020infrared}, mixed \cite{kong2021dynamic}, and syncretic \cite{wei2021syncretic}. As shown in Table \ref{tab:2}, the X-modality is generated from visible images only, ignoring the impact of infrared modality and achieving relatively low performances. Although syncretic and mixed modalities combine both visible and infrared patterns, they only utilize pixel-level information, lacking the ability to discover fine-grained and semantic patterns.

When switching back to our auxiliary modality, we observed a significant improvement in performance with a boost of 3.29\%/5.07\% in terms of rank-1/mAP on the SYSU-MM01 dataset. These results further prove that our auxiliary modality is superior to other intermediate modalities. In summary, our auxiliary modality can effectively integrate modality-related information from both visible and infrared images, while preserving strong discriminability. This promotes robust representation learning in VI-ReID.

\textbf{Effectiveness of loss design schemes.} In the modality alignment loss ($L_{ia}$), we design the modality prototype scheme to relieve the inconsistency issue and the auxiliary bridge scheme to reduce the optimization difficulty. To validate the effectiveness of these two schemes, we compare the performance of our method with or without using these two proposed schemes. As shown in Table \ref{tab:3}. 

\begin{table}[htbp]\small
  \centering
  \caption{Performance comparison of with or without using the modality prototype and auxiliary (Aux.) bridge schemes in the modality alignment loss. CMC (\%) at rank 1 and mAP (\%).}
    \begin{tabular}{cc|cc|cc}
    \hline
    \multicolumn{2}{c|}{Schemes in $L_{ma}$} & \multicolumn{2}{c|}{SYSU-MM01} & \multicolumn{2}{c}{RegDB} \bigstrut[t]\\
    \hline
    Prototype & Aux. bridge & r=1   & mAP   & r=1   & mAP \bigstrut\\
    \hline
          &       & 69.02  & 65.83  & 80.09 & 73.97 \bigstrut[t]\\
    $\checkmark$     &       & \color{blue}\textbf{74.15}  & \color{blue}\textbf{72.10}  & \color{blue}\textbf{86.73} & \color{blue}\textbf{77.24} \\
          & $\checkmark$     & 71.66  & 68.35  & 83.92 & 75.54 \\
    $\checkmark$     & $\checkmark$     & \color{red}\textbf{76.24}  & \color{red}\textbf{73.81}  & \color{red}\textbf{95.19} & \color{red}\textbf{87.15} \\
    \hline
    \end{tabular}%
  \label{tab:3}%
\end{table}%

In Table \ref{tab:3}, it is evident that both the modality prototype scheme and the auxiliary bridge scheme contribute to the VI-ReID accuracy gain. The modality prototype consistently captures global modality information, enhancing the robustness of learned modality relationships by synchronizing the alignment in each iteration; The auxiliary bridge scheme reduces the relative distance between visible and infrared features, effectively alleviating the difficulty of cross-modality distance optimization.

\subsection{Comparison with State-of-the-art Methods}
In this section, we compare our MUN with state-of-the-art works on two public datasets, as shown in Table \ref{tab:4}.

\begin{table*}[htbp]
  \centering
  \setlength{\abovecaptionskip}{0cm}
  \caption{Comparison with state-of-the-art methods on SYSU-MM01 and RegDB datasets. CMC (\%) at rank r and mAP (\%).}
  \resizebox{\linewidth}{!}{
    \begin{tabular}{c|c|ccc|ccc|ccc|ccc}
    \hline
    \multirow{3}[6]{*}{Methods} & \multirow{3}[6]{*}{Ref.} & \multicolumn{6}{c|}{SYSU-MM01}                 & \multicolumn{6}{c}{RegDB} \bigstrut\\
\cline{3-14}          &       & \multicolumn{3}{c|}{All-Search} & \multicolumn{3}{c|}{Indoor-Search} & \multicolumn{3}{c|}{Visible to Infrared} & \multicolumn{3}{c}{Infrared to Visible} \bigstrut\\
\cline{3-14}          &       & r=1   & r=10  & mAP   & r=1   & r=10  & mAP   & r=1   & r=10  & \multicolumn{1}{c|}{mAP} & r=1   & r=10  & mAP \bigstrut\\
    \hline
    Zero-padding \cite{wu2017rgb} & ICCV17 &  14.80 & 54.12 & 15.95 & 20.58& 68.38 & 26.92 & 17.75& 34.21 & \multicolumn{1}{c|}{18.90} & 16.63 & 34.68 & 17.82 \bigstrut[t]\\
    JSIA-ReID \cite{wang2020cross} & AAAI20 & 38.10  & 80.70  & 36.90  & 43.80  & 86.20  & 52.90  & 48.50  & $-$    & \multicolumn{1}{c|}{49.30} & 48.10  & $-$    & 48.90  \\
    AlignGAN \cite{wang2019rgb} & ICCV19 & 42.40  & 85.00  & 40.70  & 45.90  & 87.60  & 54.30  & 57.90  & $-$    & \multicolumn{1}{c|}{53.60 } & 56.30  & $-$    & 53.40  \\
    AGW \cite{ye2021deep} & TPAMI21 & 47.50  & 84.39  & 47.65  & 54.17  & 91.14  & 62.97  & 70.05  & 86.21  & \multicolumn{1}{c|}{67.64 } & 70.49  & 87.21  & 65.90  \\
    X-Modality \cite{li2020infrared} & AAAI20 & 49.92  & 89.79  & 50.73  & $-$    & $-$    & $-$    & 62.21  & 83.13  & \multicolumn{1}{c|}{60.18 } & $-$    & $-$    & $-$ \\
    DFLN-ViT \cite{zhao2022spatial} & TMM22 & 59.84  & 92.49  & 57.70  & 62.13  & 94.83  & 69.03  & 92.10  & 97.97  & \multicolumn{1}{c|}{82.11 } & 91.21  & \color{red}\textbf{98.20}  & 81.62  \\
    SPOT \cite{chen2022structure} & TIP22 & 65.34  & 92.73  & 62.25  & 69.42  & 96.22  & 74.63  & 80.35  & 93.48  & \multicolumn{1}{c|}{72.46 } & 79.37  & 92.79  & 72.26  \\
    FMCNet \cite{zhang2022fmcnet} & CVPR22 & 66.34  & $-$    & 62.51  & 68.15  & $-$    & 63.82  & 89.12  & $-$    & \multicolumn{1}{c|}{84.43 } & 88.23  & $-$    & 83.86  \\
    SMCL \cite{wei2021syncretic} & ICCV21 & 67.39  & 92.87  & 61.78  & 68.84  & 96.55  & 75.56  & 83.93  & $-$    & \multicolumn{1}{c|}{79.83 } & 83.05  & $-$    & 78.57  \\
    PMT \cite{lu2023learning} & AAAI23 & 67.53  & 95.36  & 64.98  & 71.66  & 96.73  & 76.52  & 84.83  & $-$    & \multicolumn{1}{c|}{76.55 } & 84.16  & $-$    & 75.13  \\
    AGW+J \cite{ye2021channel} & ICCV21 & 69.88  & 95.71  & 66.89  & 76.26  & 97.88  & 80.37  & 85.03  & 95.49  & \multicolumn{1}{c|}{79.14 } & 84.75  & 95.33  & 77.82  \\
    MPANet \cite{wu2021discover} & CVPR21 & 70.58  & 96.10  & 68.24  & 76.74  & 98.21  & \color{blue}\textbf{80.95}  & 83.70  & $-$    & \multicolumn{1}{c|}{80.90 } & 82.80  & $-$    & 80.70  \\
    CMT \cite{jiang2022cross} & ECCV22 & \color{blue}\textbf{71.88}  & \color{blue}\textbf{96.45}  & \color{blue}\textbf{68.57}  & \color{blue}\textbf{76.90}  & \color{blue}\textbf{97.68}  & 79.91  & \color{blue}\textbf{95.17}  & \color{blue}\textbf{98.82}  & \multicolumn{1}{c|}{\color{red}\textbf{87.30}} & \color{red}\textbf{91.97} & 97.92  & \color{blue}\textbf{84.46}  \bigstrut[b]\\
    \hline
    MUN (Ours)  & ICCV23 & \color{red}\textbf{76.24} & \color{red}\textbf{97.84} & \color{red}\textbf{73.81} & \color{red}\textbf{79.42} & \color{red}\textbf{98.09} & \color{red}\textbf{82.06} & \color{red}\textbf{95.19} & \color{red}\textbf{98.93} & \color{blue}\textbf{87.15}  & \color{blue}\textbf{91.86}  & \color{blue}\textbf{97.99} & \color{red}\textbf{85.01}\\
    \hline
    \end{tabular}%
    }
  \label{tab:4}%
\end{table*}%

\textbf{Comparison on SYSU-MM01 dataset.}
As illustrated in Table \ref{tab:4}, the proposed MUN achieves impressive results with 76.24\% rank-1 and 73.81\% mAP on the all-search mode of the SYSU-MM01 dataset. Compared to traditional visible-infrared representation learning methods (AGW \cite{ye2021deep}, DELN-ViT \cite{zhao2022spatial}, SPOT \cite{chen2022structure}, PMT \cite{lu2023learning}, CMT \cite{jiang2022cross}, MPANet \cite{wu2021discover}), our MUN outperforms them by a margin of at least 4.36\% rank-1 and 5.24\% mAP on the all-search mode. The reason can be attributed that the proposed visible-auxiliary-infrared learning framework can capture more identity-related knowledge across modalities and regulate the discriminative feature relationships well. Additionally, the overall performance of our approach is also superior to GAN-based methods (JSIA-ReID \cite{wang2020cross}, AlignGAN \cite{wang2019rgb}, FMCNet \cite{zhang2022fmcnet}) thanks to the powerful auxiliary modality which dynamically combines the information of visible and infrared images without introducing extra noise.

Furthermore, the proposed method significantly outperforms existing modality-unifying methods (SMCL \cite{wei2021syncretic}, X-Modality \cite{li2020infrared}) by at least 8.85\% on the rank-1 metric. This can be attributed to the fact that we not only mine the fine-grained semantic representations to generate the auxiliary modality but also decouple the extraction of specific and shared patterns in two modalities, which contribute to the dynamic generation of the auxiliary modality for relieving the changeable modality discrepancies during the training.

\textbf{Comparison on RegDB dataset.} The results on RegDB are also listed in Table \ref{tab:4}. In this dataset, image samples are spatially aligned and present less intra-class variations. Thus, the accuracy of all methods is higher than that on SYSU-MM01. The proposed MUN achieves Rank-1 of 95.19\% and mAP of 87.15\% in visible to infrared mode. Similar improvement can also be observed in the infrared to visible mode, which shows that our method obtains Rank-1 of 91.86\% and mAP of 85.01\%. This improvement can be attributed to the capacity of our method to generate a robust auxiliary modality, effectively mitigating both cross-modality and intra-modality discrepancies.

\subsection{Evaluation on Generalizability}
To verify the generalizability of MUN, we conduct experiments on two corrupted VI Re-ID datasets \cite{chen2021benchmarks}, namely SYSU-MM01-C and RegDB-C. We utilize the same corruption settings as \cite{chen2021benchmarks}, which only performs corruptions during the testing stage and randomly selects one corruption type (\eg, elastic, snow, frosted glass, etc.) and one severity level for each image in the visible gallery set. The results reported for all the compared methods are obtained using the official best settings as provided in their respective papers.

\begin{table}[t]
  \centering
  \caption{Evaluations on corrupted datasets. Each evaluation is performed 10 times to obtain the mean value. $L_{id}^a$ denotes an individual identity loss used to supervise the auxiliary modality.}
  \resizebox{\linewidth}{!}{
    \begin{tabular}{c|l|cc|cc}
    \hline
    \multirow{2}[4]{*}{Index} & \multirow{2}[4]{*}{Method} & \multicolumn{2}{c|}{SYSU-MM01-C} & \multicolumn{2}{c}{RegDB-C} \bigstrut\\
\cline{3-6}          &       & r=1   & mAP   & r=1   & mAP \bigstrut\\
    \hline
    1     & X-Modality \cite{li2020infrared} & 31.98  & 26.20  & 37.26  & 35.97  \bigstrut[t]\\
    2     & CIL \cite{chen2021benchmarks} & 36.95  & 35.92  & \color{blue}\textbf{52.25}  & \color{blue}\textbf{49.76}  \\
    3     & SMCL \cite{wei2021syncretic} & 37.08  & 36.12  & 51.93  & 49.22  \\
    4     & AGW+J \cite{ye2021channel} & \color{blue}\textbf{40.09}  & \color{blue}\textbf{37.86}  & 51.53  & 49.04  \bigstrut[b]\\
    \hline
    5     & B     & 25.92  & 23.13  & 32.05  & 29.64  \bigstrut[t]\\
    6     & +Aux.+$L_{id}^a$   & 30.85  & 26.54  & 36.40  & 31.21  \\
    7     & +Aux.+$L_{idc}$ & 31.12  & 27.44  & 38.13  & 32.29  \\
    8     & +Aux.+$L_{idc}$+$L_{ia}$ & 36.78  & 32.32  & 45.44  & 42.89  \\
    9     & +Aux.+$L_{idc}$+$L_{ia}$+$L_{ma}$ & \color{red}\textbf{41.17}  & \color{red}\textbf{38.63}  & \color{red}\textbf{52.69}  & \color{red}\textbf{50.18}\\
    \hline
    \end{tabular}%
    }
  \label{tab:5}%
  \vspace{-1.0em}
\end{table}%

Table \ref{tab:5} shows that the performance of baseline (B) is relatively lower than that of existing SOTAs under data corruption scenarios. However, by introducing the auxiliary modality to bridge the gap between visible and infrared modalities (Index 6), the accuracy is significantly improved. Furthermore, by incorporating the proposed identity alignment loss ($L_{ia}$) and modality alignment loss ($L_{ma}$) to refine the learned cross-modality feature relationships, the rank-1 and mAP accuracies experience a substantial enhancement of 15.25\% and 15.5\% respectively on the SYSU-MM01-C dataset. This outperforms the current SOTAs by a remarkable margin. Specifically, our MUN method exceeds the highly robust AGW+J \cite{ye2021channel} by 1.08\% in rank-1 and 0.77\% in mAP. The experiments on two corrupted datasets verify the strong generalizability and robustness of the proposed MUN, which can consistently learn modality-shared patterns and regulate stable feature relationships under corrupted data scenarios.

\subsection{Visualization}

\begin{figure}[t]
\centering
\includegraphics[width=0.99\linewidth]{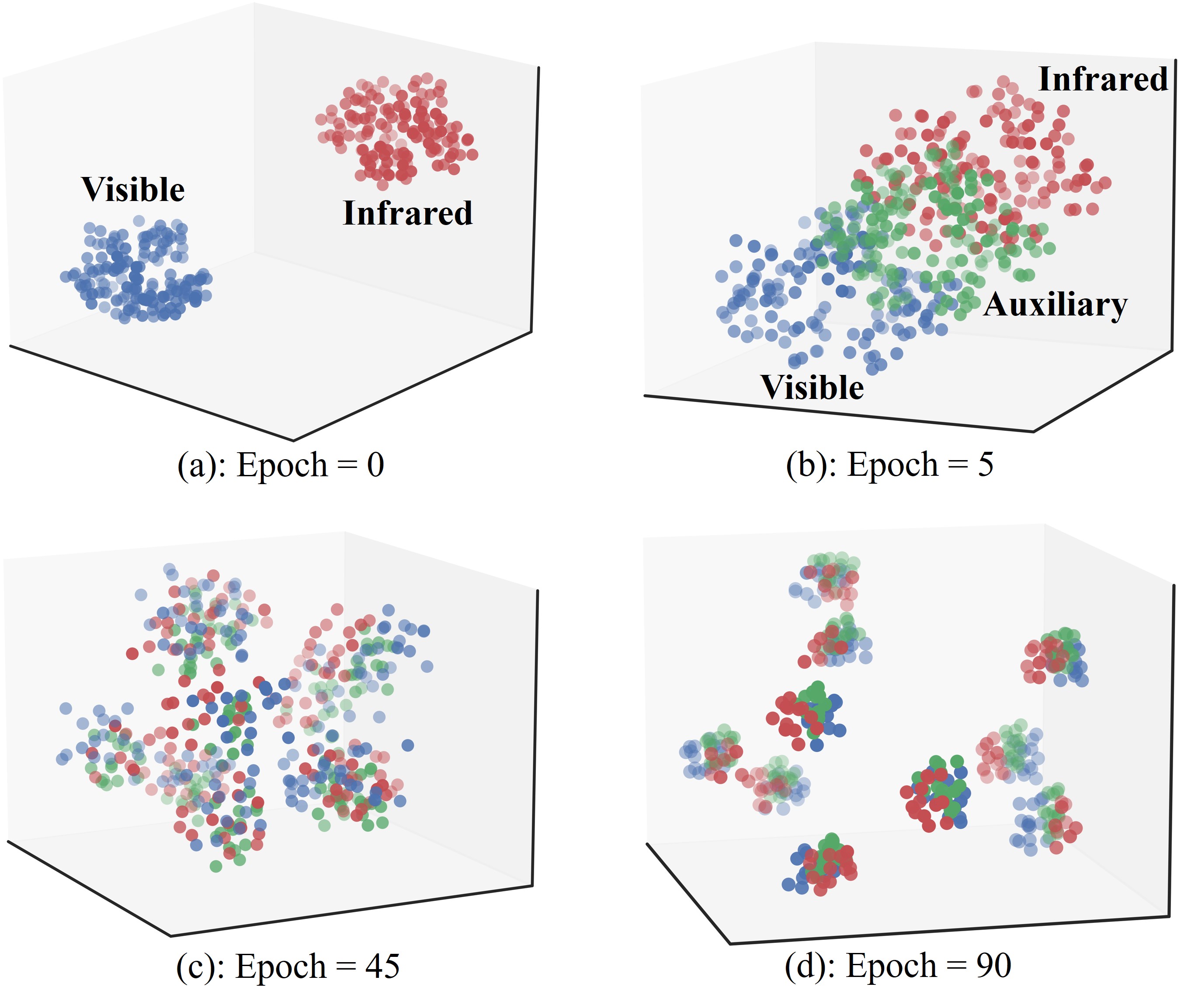}
\setlength{\abovecaptionskip}{0cm}
\caption{Distributions of learned visible, infrared and auxiliary features during the training. All the samples are randomly selected in the SYSU-MM01 dataset. The 3D visualization is made by T-SNE \cite{van2008visualizing}. Please view in colour and zoom in.}
\label{fig:4}
\end{figure}

\textbf{Distribution Visualization.} To visually demonstrate the effectiveness of MUN, we randomly select 10 identities from the SYSU-MM01 dataset and visualize their feature distributions during the training. The visualization results are presented in Figure \ref{fig:4}.

At the onset of training (epoch 0 in Figure \ref{fig:4} (a)), significant modality disparities arise between visible (depicted by blue dots) and infrared images (depicted by red dots), rendering cross-modality matching unfeasible. As the training progresses, the proposed auxiliary modality (depicted by green dots) serves as a bridge to connect the visible and infrared modalities in the common feature space. We can observe that the learned visible and infrared features show a convergence trend, and the modality discrepancies are gradually eliminated at epoch 5 in Figure \ref{fig:4} (b). Afterwards, the network learns identity-aware patterns by regulating smaller intra-class distances and larger inter-class distances at epoch 45. Finally, in Figure \ref{fig:4} (d), all the learned features are well grouped into their respective identity centers, demonstrating powerful discriminability under cross-modality scenes, which proves the effectiveness of MUN in learning robustness and identity-aware features.

\begin{figure}[htpb]
\centering
\includegraphics[width=0.99\linewidth]{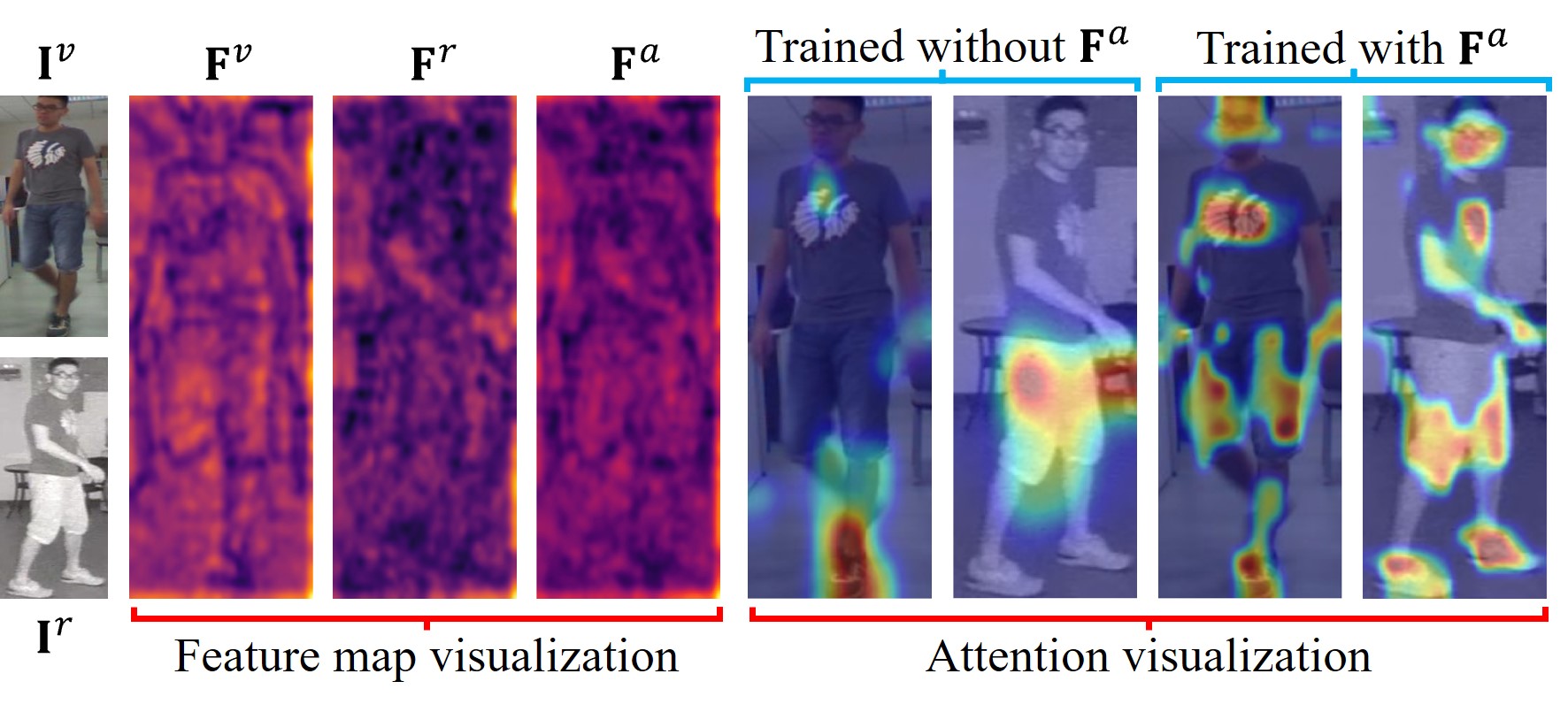}
\setlength{\abovecaptionskip}{0cm}
\caption{Visualization of learned feature maps and attention maps. Given the input visible image $\textbf{I}^v$ and infrared image $\textbf{I}^r$, we visualize the corresponding learned visible feature $\textbf{F}^v$, infrared feature $\textbf{F}^r$, and the generated auxiliary feature $\textbf{F}^a$. It is obvious that the auxiliary feature $\textbf{F}^a$ can preserve most of the modality-shared patterns, including body shape and structure. The attention visualizations on the input images are made by Grad-CAM \cite{selvaraju2017grad}. Please view in colour and zoom in.}
\label{fig:5}
\end{figure}

\textbf{Pattern Visualization.} In order to further illustrate the effectiveness of our auxiliary modality, we select one visible and one infrared image with the same identity in the SYSU-MM01 dataset to visualize the corresponding learned feature maps and attention maps via Grad-CAM \cite{selvaraju2017grad}.

The visualization results are shown in Figure \ref{fig:5}. It is clear that the visible ($\textbf{F}^v$) and infrared ($\textbf{F}^r$) feature maps extracted from the backbone have different patterns and structures, which makes it difficult to perform cross-modality matching. Notably, the proposed auxiliary generator can dynamically reconstruct and align the modality-shared spatial patterns using multiple transposed convolutions. This makes our auxiliary feature ($\textbf{F}^a$ in Figure \ref{fig:5}) preserves most of the shared patterns between $\textbf{F}^v$ and $\textbf{F}^r$ without corrupting the structure information of person bodies. In addition, the attention visualization in Figure \ref{fig:5} further indicates that the proposed auxiliary modality plays a critical role in helping the network learn modality-shared representations.

\section{Conclusion}
This paper proposes a novel Modality Unifying Network to jointly explore the robust auxiliary modality and generalize cross-modality feature relationships for VI-ReID. The auxiliary modality is generated by combining the cross-modality learner and the intra-modality learner, enabling the dynamic extraction of modality-specific patterns from multiple receptive fields and feature scales. This approach empowers our auxiliary modality to effectively alleviate both cross-modality and intra-modality discrepancies. Moreover, we propose identity alignment loss and modality alignment loss to regulate discriminative feature relationships in multi-modality tasks. Extensive experiments on public datasets demonstrate the effectiveness and generalizability of our MUN as well as each proposed component.

\noindent\textbf{Acknowledgements.} This work is supported by the Academy of Finland for Academy Professor project EmotionAI (Grants No. 336116, 345122), the University of Oulu \& The Academy of Finland Profi 7 (Grant No. 352788), and the National Natural Science Foundation of China (Grant No. 61802058, 61911530397). We appreciate the professional and cost-effective GPU computing service provided by www.AutoDL.com.  

{\small
\bibliographystyle{ieee_fullname}
\bibliography{egpaper_final}

\begin{thebibliography}{10}\itemsep=-1pt

\bibitem{chen2022structure}
Cuiqun Chen, Mang Ye, Meibin Qi, Jingjing Wu, Jianguo Jiang, and Chia-Wen Lin.
\newblock Structure-aware positional transformer for visible-infrared person re-identification.
\newblock {\em IEEE Transactions on Image Processing}, 31:2352--2364, 2022.

\bibitem{chen2021benchmarks}
Minghui Chen, Zhiqiang Wang, and Feng Zheng.
\newblock Benchmarks for corruption invariant person re-identification.
\newblock {\em arXiv preprint arXiv:2111.00880}, 2021.

\bibitem{dai2018cross}
Pingyang Dai, Rongrong Ji, Haibin Wang, Qiong Wu, and Yuyu Huang.
\newblock Cross-modality person re-identification with generative adversarial training.
\newblock In {\em IJCAI}, volume~1, page~6, 2018.

\bibitem{he2016deep}
Kaiming He, Xiangyu Zhang, Shaoqing Ren, and Jian Sun.
\newblock Deep residual learning for image recognition.
\newblock In {\em Proceedings of the IEEE conference on computer vision and pattern recognition}, pages 770--778, 2016.

\bibitem{jambigi2021mmd}
Chaitra Jambigi, Ruchit Rawal, and Anirban Chakraborty.
\newblock Mmd-reid: A simple but effective solution for visible-thermal person reid.
\newblock {\em arXiv preprint arXiv:2111.05059}, 2021.

\bibitem{jiang2022cross}
Kongzhu Jiang, Tianzhu Zhang, Xiang Liu, Bingqiao Qian, Yongdong Zhang, and Feng Wu.
\newblock Cross-modality transformer for visible-infrared person re-identification.
\newblock In {\em Computer Vision--ECCV 2022: 17th European Conference, Tel Aviv, Israel, October 23--27, 2022, Proceedings, Part XIV}, pages 480--496. Springer, 2022.

\bibitem{kong2021dynamic}
Jun Kong, Qibin He, Min Jiang, and Tianshan Liu.
\newblock Dynamic center aggregation loss with mixed modality for visible-infrared person re-identification.
\newblock {\em IEEE Signal Processing Letters}, 28:2003--2007, 2021.

\bibitem{leng2019survey}
Qingming Leng, Mang Ye, and Qi Tian.
\newblock A survey of open-world person re-identification.
\newblock {\em IEEE Transactions on Circuits and Systems for Video Technology}, 30(4):1092--1108, 2019.

\bibitem{li2020infrared}
Diangang Li, Xing Wei, Xiaopeng Hong, and Yihong Gong.
\newblock Infrared-visible cross-modal person re-identification with an x modality.
\newblock In {\em Proceedings of the AAAI Conference on Artificial Intelligence}, volume~34, pages 4610--4617, 2020.

\bibitem{liao2022graph}
Shengcai Liao and Ling Shao.
\newblock Graph sampling based deep metric learning for generalizable person re-identification.
\newblock In {\em Proceedings of the IEEE/CVF Conference on Computer Vision and Pattern Recognition}, pages 7359--7368, 2022.

\bibitem{liu2022survey}
Minjie Liu, Jiaqi Zhao, Yong Zhou, Hancheng Zhu, Rui Yao, and Ying Chen.
\newblock Survey for person re-identification based on coarse-to-fine feature learning.
\newblock {\em Multimedia Tools and Applications}, pages 1--35, 2022.

\bibitem{liu2022convnet}
Zhuang Liu, Hanzi Mao, Chao-Yuan Wu, Christoph Feichtenhofer, Trevor Darrell, and Saining Xie.
\newblock A convnet for the 2020s.
\newblock In {\em Proceedings of the IEEE/CVF Conference on Computer Vision and Pattern Recognition}, pages 11976--11986, 2022.

\bibitem{loshchilov2018fixing}
Ilya Loshchilov and Frank Hutter.
\newblock Fixing weight decay regularization in adam.
\newblock 2018.

\bibitem{lu2023learning}
Hu Lu, Xuezhang Zou, and Pingping Zhang.
\newblock Learning progressive modality-shared transformers for effective visible-infrared person re-identification.
\newblock In {\em Proceedings of the AAAI Conference on Artificial Intelligence}, volume~37, pages 1835--1843, 2023.

\bibitem{luo2019bag}
Hao Luo, Youzhi Gu, Xingyu Liao, Shenqi Lai, and Wei Jiang.
\newblock Bag of tricks and a strong baseline for deep person re-identification.
\newblock In {\em Proceedings of the IEEE/CVF conference on computer vision and pattern recognition workshops}, pages 0--0, 2019.

\bibitem{luo2019strong}
Hao Luo, Wei Jiang, Youzhi Gu, Fuxu Liu, Xingyu Liao, Shenqi Lai, and Jianyang Gu.
\newblock A strong baseline and batch normalization neck for deep person re-identification.
\newblock {\em IEEE Transactions on Multimedia}, 22(10):2597--2609, 2019.

\bibitem{mathur2020brief}
Neha Mathur, Shruti Mathur, Divya Mathur, and Pankaj Dadheech.
\newblock A brief survey of deep learning techniques for person re-identification.
\newblock In {\em 2020 3rd International Conference on Emerging Technologies in Computer Engineering: Machine Learning and Internet of Things (ICETCE)}, pages 129--138. IEEE, 2020.

\bibitem{nguyen2017person}
Dat~Tien Nguyen, Hyung~Gil Hong, Ki~Wan Kim, and Kang~Ryoung Park.
\newblock Person recognition system based on a combination of body images from visible light and thermal cameras.
\newblock {\em Sensors}, 17(3):605, 2017.

\bibitem{ning2020feature}
Xin Ning, Ke Gong, Weijun Li, Liping Zhang, Xiao Bai, and Shengwei Tian.
\newblock Feature refinement and filter network for person re-identification.
\newblock {\em IEEE Transactions on Circuits and Systems for Video Technology}, 31(9):3391--3402, 2020.

\bibitem{park2021learning}
Hyunjong Park, Sanghoon Lee, Junghyup Lee, and Bumsub Ham.
\newblock Learning by aligning: Visible-infrared person re-identification using cross-modal correspondences.
\newblock In {\em Proceedings of the IEEE/CVF International Conference on Computer Vision}, pages 12046--12055, 2021.

\bibitem{radenovic2018fine}
Filip Radenovi{\'c}, Giorgos Tolias, and Ond{\v{r}}ej Chum.
\newblock Fine-tuning cnn image retrieval with no human annotation.
\newblock {\em IEEE transactions on pattern analysis and machine intelligence}, 41(7):1655--1668, 2018.

\bibitem{selvaraju2017grad}
Ramprasaath~R Selvaraju, Michael Cogswell, Abhishek Das, Ramakrishna Vedantam, Devi Parikh, and Dhruv Batra.
\newblock Grad-cam: Visual explanations from deep networks via gradient-based localization.
\newblock In {\em Proceedings of the IEEE international conference on computer vision}, pages 618--626, 2017.

\bibitem{van2008visualizing}
Laurens Van~der Maaten and Geoffrey Hinton.
\newblock Visualizing data using t-sne.
\newblock {\em Journal of machine learning research}, 9(11), 2008.

\bibitem{wang2019rgb}
Guan'an Wang, Tianzhu Zhang, Jian Cheng, Si Liu, Yang Yang, and Zengguang Hou.
\newblock Rgb-infrared cross-modality person re-identification via joint pixel and feature alignment.
\newblock In {\em Proceedings of the IEEE/CVF International Conference on Computer Vision}, pages 3623--3632, 2019.

\bibitem{wang2020cross}
Guan-An Wang, Tianzhu Zhang, Yang Yang, Jian Cheng, Jianlong Chang, Xu Liang, and Zeng-Guang Hou.
\newblock Cross-modality paired-images generation for rgb-infrared person re-identification.
\newblock In {\em Proceedings of the AAAI Conference on Artificial Intelligence}, volume~34, pages 12144--12151, 2020.

\bibitem{wang2019learning}
Zhixiang Wang, Zheng Wang, Yinqiang Zheng, Yung-Yu Chuang, and Shin'ichi Satoh.
\newblock Learning to reduce dual-level discrepancy for infrared-visible person re-identification.
\newblock In {\em Proceedings of the IEEE/CVF Conference on Computer Vision and Pattern Recognition}, pages 618--626, 2019.

\bibitem{wei2021syncretic}
Ziyu Wei, Xi Yang, Nannan Wang, and Xinbo Gao.
\newblock Syncretic modality collaborative learning for visible infrared person re-identification.
\newblock In {\em Proceedings of the IEEE/CVF International Conference on Computer Vision}, pages 225--234, 2021.

\bibitem{wu2017rgb}
Ancong Wu, Wei-Shi Zheng, Hong-Xing Yu, Shaogang Gong, and Jianhuang Lai.
\newblock Rgb-infrared cross-modality person re-identification.
\newblock In {\em Proceedings of the IEEE international conference on computer vision}, pages 5380--5389, 2017.

\bibitem{wu2021discover}
Qiong Wu, Pingyang Dai, Jie Chen, Chia-Wen Lin, Yongjian Wu, Feiyue Huang, Bineng Zhong, and Rongrong Ji.
\newblock Discover cross-modality nuances for visible-infrared person re-identification.
\newblock In {\em Proceedings of the IEEE/CVF Conference on Computer Vision and Pattern Recognition}, pages 4330--4339, 2021.

\bibitem{yan2021beyond}
Cheng Yan, Guansong Pang, Xiao Bai, Changhong Liu, Xin Ning, Lin Gu, and Jun Zhou.
\newblock Beyond triplet loss: person re-identification with fine-grained difference-aware pairwise loss.
\newblock {\em IEEE Transactions on Multimedia}, 24:1665--1677, 2021.

\bibitem{ye2021channel}
Mang Ye, Weijian Ruan, Bo Du, and Mike~Zheng Shou.
\newblock Channel augmented joint learning for visible-infrared recognition.
\newblock In {\em Proceedings of the IEEE/CVF International Conference on Computer Vision}, pages 13567--13576, 2021.

\bibitem{ye2020dynamic}
Mang Ye, Jianbing Shen, David J~Crandall, Ling Shao, and Jiebo Luo.
\newblock Dynamic dual-attentive aggregation learning for visible-infrared person re-identification.
\newblock In {\em European Conference on Computer Vision}, pages 229--247. Springer, 2020.

\bibitem{ye2021deep}
Mang Ye, Jianbing Shen, Gaojie Lin, Tao Xiang, Ling Shao, and Steven~CH Hoi.
\newblock Deep learning for person re-identification: A survey and outlook.
\newblock {\em IEEE transactions on pattern analysis and machine intelligence}, 44(6):2872--2893, 2021.

\bibitem{ye2018visible}
Mang Ye, Zheng Wang, Xiangyuan Lan, and Pong~C Yuen.
\newblock Visible thermal person re-identification via dual-constrained top-ranking.
\newblock In {\em IJCAI}, volume~1, page~2, 2018.

\bibitem{zeng2020hierarchical}
Kaiwei Zeng, Munan Ning, Yaohua Wang, and Yang Guo.
\newblock Hierarchical clustering with hard-batch triplet loss for person re-identification.
\newblock In {\em Proceedings of the IEEE/CVF Conference on Computer Vision and Pattern Recognition}, pages 13657--13665, 2020.

\bibitem{zhang2022fmcnet}
Qiang Zhang, Changzhou Lai, Jianan Liu, Nianchang Huang, and Jungong Han.
\newblock Fmcnet: Feature-level modality compensation for visible-infrared person re-identification.
\newblock In {\em Proceedings of the IEEE/CVF Conference on Computer Vision and Pattern Recognition}, pages 7349--7358, 2022.

\bibitem{zhang2022cross}
Sen Zhang, Zhaowei Shang, Mingliang Zhou, Yingxin Wang, and Guoliang Sun.
\newblock Cross-modal identity correlation mining for visible-thermal person re-identification.
\newblock {\em Multimedia Tools and Applications}, pages 1--14, 2022.

\bibitem{zhao2022spatial}
Jiaqi Zhao, Hanzheng Wang, Yong Zhou, Rui Yao, Silin Chen, and Abdulmotaleb El~Saddik.
\newblock Spatial-channel enhanced transformer for visible-infrared person re-identification.
\newblock {\em IEEE Transactions on Multimedia}, 2022.

\bibitem{zheng2019joint}
Zhedong Zheng, Xiaodong Yang, Zhiding Yu, Liang Zheng, Yi Yang, and Jan Kautz.
\newblock Joint discriminative and generative learning for person re-identification.
\newblock In {\em proceedings of the IEEE/CVF conference on computer vision and pattern recognition}, pages 2138--2147, 2019.

\bibitem{zhu2020aware}
Zhihui Zhu, Xinyang Jiang, Feng Zheng, Xiaowei Guo, Feiyue Huang, Xing Sun, and Weishi Zheng.
\newblock Aware loss with angular regularization for person re-identification.
\newblock In {\em Proceedings of the AAAI conference on artificial intelligence}, volume~34, pages 13114--13121, 2020.

\end{thebibliography}
}

\end{document}